\documentclass[runningheads]{llncs}

\usepackage{eccv}
\usepackage{eccvabbrv}
\usepackage{graphicx}
\usepackage{booktabs}
\usepackage[accsupp]{axessibility} 

\usepackage[pagebackref,breaklinks,colorlinks,citecolor=eccvblue]{hyperref}
\usepackage{orcidlink}

\begin{document}

\title{Improving Post-Earthquake Crack Detection using Semi-Synthetic Generated Images} 
\titlerunning{Improving Post-Earthquake Crack Detection}

\author{
Piercarlo Dondi\inst{1}\orcidlink{0000-0002-0624-073X} \and 
Alessio Gullotti\inst{1}\orcidlink{0000-0003-2926-6796} \and
Michele Inchingolo\inst{1}\orcidlink{0009-0002-4795-3047} \and
Ilaria Senaldi\inst{2}\orcidlink{0000-0001-8858-9806} \and
Chiara Casarotti \inst{2}\orcidlink{0000-0003-1801-6900} \and 
Luca Lombardi\inst{1}\orcidlink{0000-0002-5457-7556} \and
Marco Piastra\inst{1}\orcidlink{0000-0003-2556-5254}  
}

\authorrunning{P. Dondi et al.}

\institute{
Department of Electrical, Computer and Biomedical Engineering, \\University of Pavia, Via Ferrata 5, 27100 Pavia, Italy\\
\email{\{piercarlo.dondi, luca.lombardi, marco.piastra\}@unipv.it; \{alessio.gullotti01, michele.inchingolo01\}@universitadipavia.it}
\and
EUCENTRE Foundation, European Centre for Training and Research in Earthquake Engineering, Via Ferrata 1, 27100 Pavia, Italy\\
\email{\{ilaria.senaldi, chiara.casarotti\}@eucentre.it}
}

\maketitle

\begin{abstract}
Following an earthquake, it is vital to quickly evaluate the safety of the impacted areas. Damage detection systems, powered by computer vision and deep learning, can assist experts in this endeavor. However, the lack of extensive, labeled datasets poses a challenge to the development of these systems. In this study, we introduce a technique for generating semi-synthetic images to be used as data augmentation during the training of a damage detection system. We specifically aim to generate images of cracks, which are a prevalent and indicative form of damage. The central concept is to employ parametric meta-annotations to guide the process of generating cracks on 3D models of real-word structures. The governing parameters of these meta-annotations can be adjusted iteratively to yield images that are optimally suited for improving detectors' performance. Comparative evaluations demonstrated that a crack detection system trained with a combination of real and semi-synthetic images outperforms a system trained on real images alone.

\end{abstract}

\section{Introduction}
\label{sec:intro}
After a catastrophic event, such as an earthquake, it is crucial to perform a rapid structural safety assessment of the endangered areas. Nowadays, common approaches involve the direct inspection by human experts of photos and videos acquired in the field, generally using Unmanned Aerial Systems (UAS) \cite{Mandirola2022,Seo2018}. Such an examination is a slow process that can be sped up by Computer Vision (CV) and Deep Learning (DL) solutions \cite{Sony2021}. Automatic damage detection algorithms are particularly helpful for this task, since they can provide a fast preliminary screening of images and videos, thus reducing the amount of data that experts need to analyze.

However, the creation of damage detectors that utilize Deep Convolutional Neural Networks (DCNN) usually necessitates a substantial amount of labeled data. Indeed, one of the primary challenges in building such systems is the limited availability of large, open-source datasets of annotated images. Furthermore, many of the currently available datasets consist of images derived from routine structural monitoring. Such datasets typically represent less severe damage levels compared to what might be observed in the following of an earthquake \cite{bianchi2022,Yang2022}.

Data augmentation approaches have been proposed in literature, either based on standard image transformations \cite{Polovnikov2021, Zhao2024}, generative models \cite{Jain2022,Branikas2023,Li2024} or custom techniques for specific types of structures \cite{Boikov2021, Xu2023}. Nevertheless, to the best of authors' knowledge, no data augmentation approach exists for a post-earthquake scenario, in which the images to be generated may include diverse types of severely damaged structures, framed at different distance and under different meteorological conditions.

In the present work, we propose a method for creating semi-synthetic images whereby computer-generated damage is applied on photorealistic real-world 3D models. We focus here on the generation of \textit{cracks}, which is one of the most common and possibly more revealing types of damage in the domain considered. Real-world cracks can vary widely in shape, size, and location, depending on the level of severity and the type of structure affected. To handle such complexity and variety, producing realistic outcomes, we designed a set of parametric \textit{meta-annotations} that could govern the random generation of crack instances within a specific set of constraints defined by human experts.
The overall procedure starts from 3D models of real-world buildings and bridges obtained by photogrammetric elaboration of images acquired by UAS. Meta-annotations are then added and defined in appropriate locations. Finally, images are rendered by specifying virtual camera paths over the 3D models, with varying lights and ambient conditions. Following this procedure, it is possible to generate a large number of images, also simulating different scenarios and levels of damage. We believe that this approach could produce more realistic outcomes than those achievable from fully artificial CAD models, by also harnessing inspection by UAS flights that are possibly simpler and less expensive to acquire. In comparison to the usage of generative AI, such as diffusion models, the proposed method is intended to allow a higher degree of expert-based control over the generation and variety of crack instances than that achievable by generative network conditioning. 

By design, semi-synthetic images produced in this way are intended as an augmentation to datasets of real images, and not as a complete replacement. In addition, the meta-annotation parameters can be tuned incrementally to match the progress of the training process. The working hypothesis is that the iterative application of this tunable augmentation strategy could improve the performance of a DCNN-based crack detector. To prove it, we performed comparative experiments using various off-the-shelf YOLO DCNN architectures. In these experiments, three detectors were trained: one using real images only, another using semi-synthetic images only, and the last one using a combination of real and semi-synthetic images. All these combinations were evaluated on the same test set of real images. As the basis for both training and testing, we used the IDEA dataset by EUCENTRE Foundation, first introduced in \cite{Dondi2023}. 

The remainder of the paper is structured as follows: Section \ref{sec:sota} provides a quick overview of state-of-the-art in the field; Section \ref{sec:method} describes the proposed approach; Section \ref{sec:case_study} illustrates the case study; Section \ref{sec:results} presents the achieved results; finally, Section \ref{sec:conclusions} draws the conclusions and proposes possible future developments.

\section{Previous Works}
\label{sec:sota}
The assessment of structural safety is a well-researched topic, with numerous machine learning and deep learning solutions proposed in scientific literature \cite{Flah2021,Sony2021}. These solutions primarily focus on detecting damage, particularly cracks. \cite{Hsieh2020, Yang2022}. The approaches can be categorized into three main Computer Vision tasks: classification, segmentation and object detection. 

Among the several crack classification methods proposed, notable examples include: the DCNN designed by \cite{Cha2017} for recognizing cracks in concrete structures; the SVM-based model used by \cite{Davoudi2018a} to classify crack pattern in reinforced concrete beams and slabs; the DCNN architecture employed by \cite{Dung2019} to find cracks in gusset joints in steel bridges; the DCNN proposed by \cite{Xu2019} specialized for classifying cracks on bridges. Transfer learning methods, such as the one proposed by \cite{Yang2020}, have been explored, too.

Regarding crack segmentation, various networks have been tested, including those by \cite{Zheng2020} for concreted buildings, and \cite{Dais2021} for masonry walls. A DCNN-based semantic segmentation approach was proposed by \cite{Yang2018} for identifying and measuring cracks on pavements and concrete walls. Custom crack segmentation networks have also been developed, such as CiNet \cite{Ye2019}, for concrete beams, and SDDnet \cite{Kang2020}, for buildings. Some researchers have treated the crack segmentation problem as an anomaly detection task, using  convolutional autoenconders \cite{Chow2020}. Hybrid solutions combining detection and segmentation have also been investigated, such as those for concrete cracks \cite{Choi2020} and road cracks \cite{Nguyen2021}. Recently, a Bayesian approach was proposed by \cite{Canchila2024} to optimize the hyperparameters of a DCNN-based crack segmentation algorithm.

It should be noted that both classification and segmentation methods primarily operate on close-up images or small patches of larger images. As a result, they are better suited for monitoring purposes where the target areas are predefined. In contrast, post-earthquake surveys typically occur at a mid-distance from the structures being inspected, to grant rapid information acquisition. Object detection algorithms are better suited for this scenario. Meaningful examples of crack detection methods include: the road damage detection for smarthphone proposed by \cite{Maeda2018}; the Meta-Learning Convolutional Neural Architectures proposed by \cite{Mundt2019}, designed to identify multiple classes of damage on concrete bridges; the End-to-end Defect Detection Network (EDDN) proposed by \cite{Lv2020}, designed for metal defect detection; and the crack detector for dams proposed by \cite{Xu2023}.

In terms of data availability, large datasets typically consist of image-wise classification labels (e.g., PEER Hub ImageNet \cite{Gao2020}) or close-up patches depicting specific damaged structures  (e.g., concrete walls and floors \cite{Ozgene2018} or concrete buildings \cite{Xu2019}). Moreover, datasets that include multiple damage classes and bounding box annotations are relatively scarce, containing only a few thousand images (e.g., CODEBRIM \cite{Mundt2019}). 

Many researchers have acknowledged this data scarcity and proposed various data augmentation strategies. Some have employed traditional image processing techniques, such as geometrical and color transformation \cite{Polovnikov2021} or random crack displacements \cite{Zhao2024}. Others have explored synthetic image generation methods, with Generative Adversarial Networks (GAN) being the most common approach \cite{Jain2022,Branikas2023,Li2024}. Customized CNN \cite{Kim2023} and 3D modeling techniques \cite{Boikov2021} have been tested, too. However, most of these solutions focus on generating close-up images of damage. Only a few methods address mid-distance image generation, such as the dam crack generator designed by \cite{Xu2023}. 

\section{Methodology}
\label{sec:method}

\subsection{Semi-Synthetic Image Generation Procedure}
\label{sec:image_generation}
The overall procedure presented here has the objective to create a large number of semi-synthetic annotated images to be used as data augmentation during the training of a DCNN-based crack detector. 

\begin{figure}[htb]
    \centering
    \includegraphics[width=\linewidth]{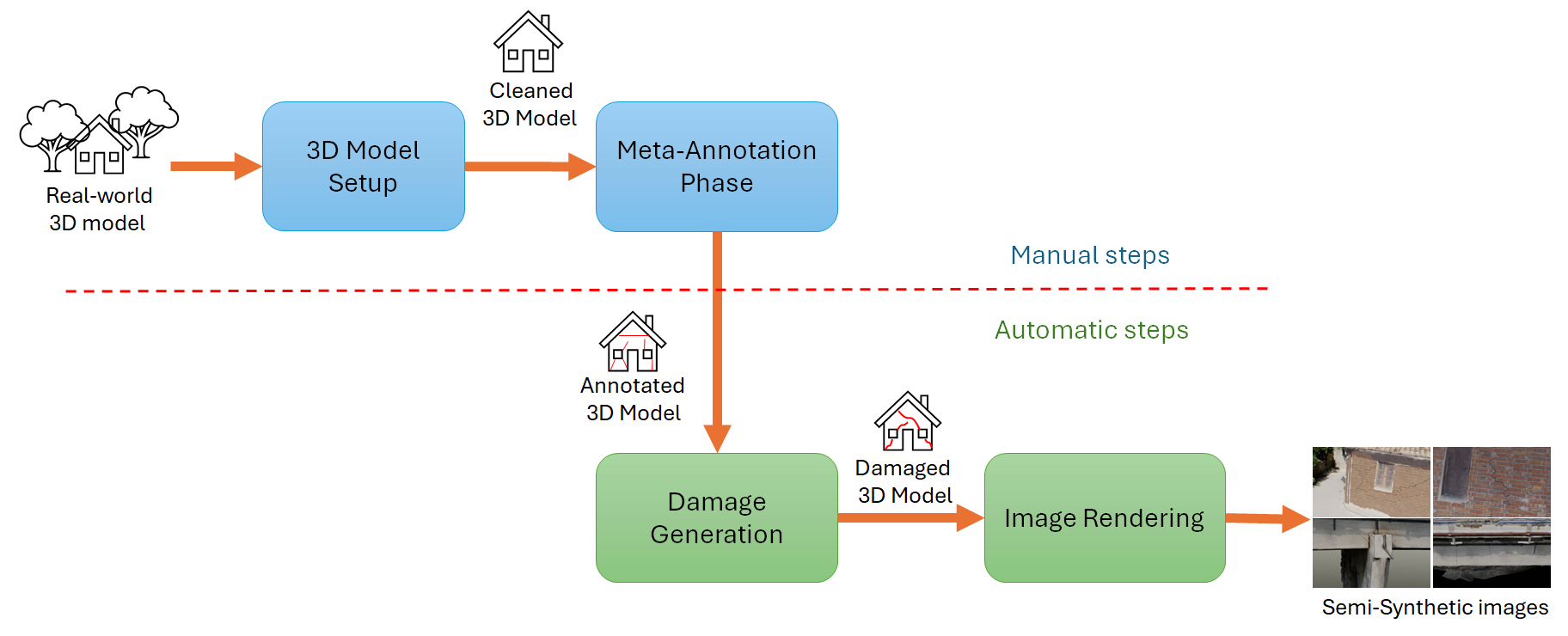}
    \caption{Workflow of the proposed semi-synthetic image generation procedure.}
    \label{fig:workflow}
\end{figure}

The proposed approach is designed to work with different types of civil structures, including masonry and concrete buildings and bridges. 3D models of real-world, undamaged structures obtained via photogrammetry will be used as input. 
The core idea is to use a set of parametric meta-annotations, manually placed by human experts, from which cracks are automatically generated. This meta-annotation process has been designed in strict cooperation with structural engineers, to guarantee the realism of the outcome. This is the major difference with respect to similar approaches described in literature \cite{Xu2023}, in which crack patterns extracted from real images are modified and applied to real-world 3D models of structures. In our opinion, the proposed method allows a higher degree of control and greater variability in the image generation process, as well as the applicability to different types of surfaces and structures. 

Figure \ref{fig:workflow} describes the workflow of the entire procedure. All steps, detailed in the following, were implemented with Blender (version 3.6 LTS) \cite{Blender} by means of a series of node-based workflows and add-on Python scripts. 

\subsubsection{3D Model Setup}
\begin{figure}[htb]
    \centering
    \includegraphics[width=0.9\linewidth]{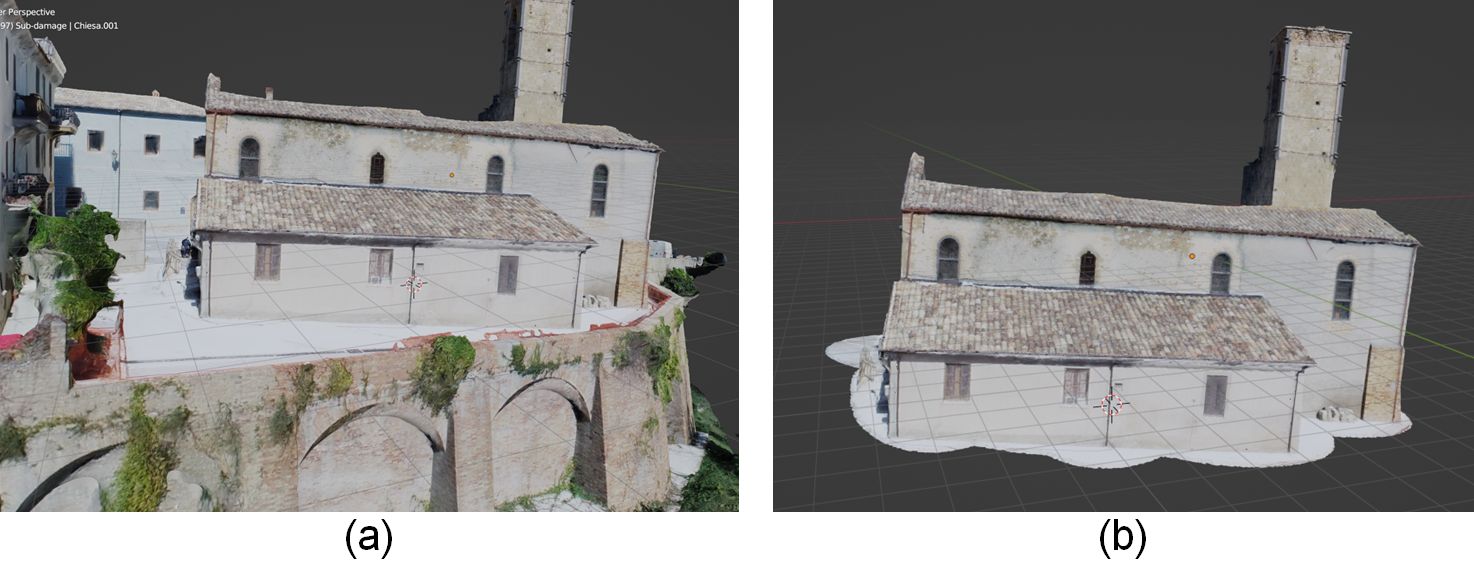}
     \caption{Initial steps of the semi-synthetic image generation procedure: (a) Loading and examination of the 3D model; (b) selection of a relevant component to be damaged.}
    \label{fig:component}
\end{figure}

The first step of the procedure involves the examination and preparation of the 3D model in input. Since a 3D model may contain multiple buildings and/or structures (e.g., it could be a large scan of an entire neighborhood), specific \textit{components} are selected, supposedly with the higher quality, and considered relevant for the following stages (Fig. \ref{fig:component}). If necessary, 3D mesh defects in the chosen components can be fixed at this stage before further processing. The remaining parts of the model will be used as background for the final rendering. 

To improve the variability of the outcomes, external textures of the components can be modified or substituted as well, for instance to turn a plastered wall into one with exposed masonry.

\begin{figure}[htb]
    \centering
    \includegraphics[width=0.85\linewidth]{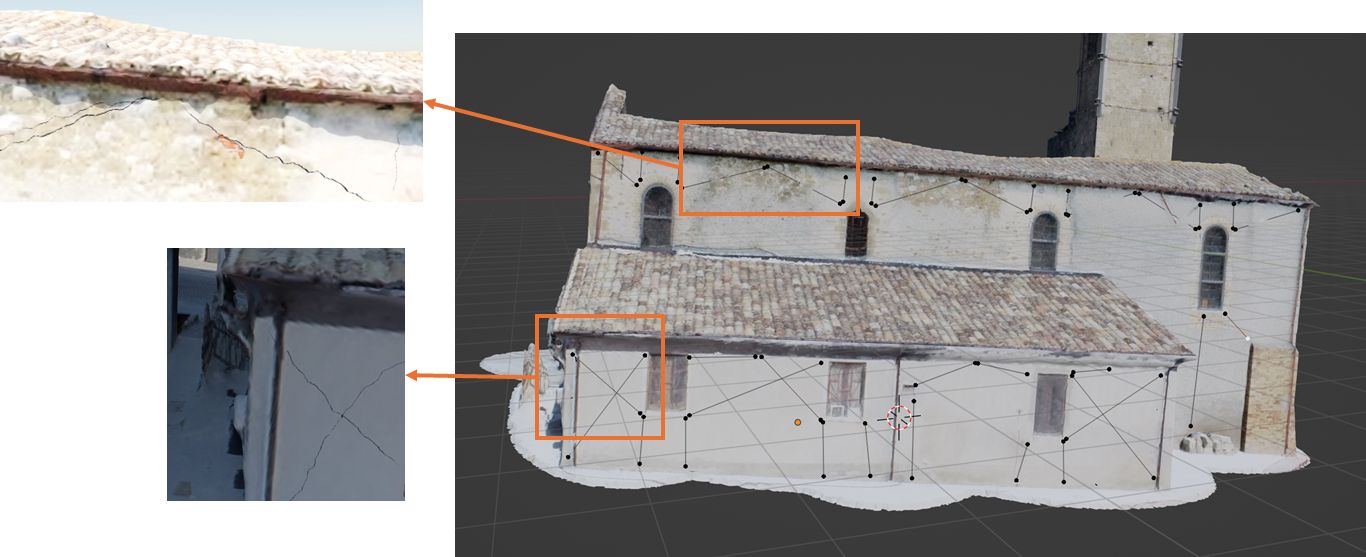}
    \caption{Example of meta-annotations (black lines) and some details of two sample renderings showing cracks thus obtained.}
    \label{fig:annotation_results}
\end{figure}

\subsubsection{Meta-Annotation Phase}
Meta-annotations are intended to define the positioning and general constraints that will govern the generation of cracks. The basic graphical primitive for defining a meta-annotation consists in a couple of points connected by a line, as shown in Fig. \ref{fig:annotation_results}. Meta-annotations are manually defined by a structural engineering expert who places them in specific positions of the selected component. 

\begin{figure}[htb]
    \centering
    \includegraphics[width=0.9\linewidth]{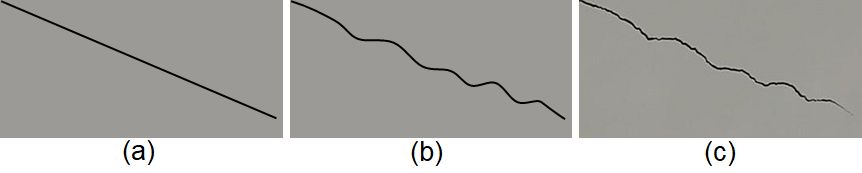}
    \caption{Example of the effect of the roughness parameter: (a) initial meta-annotation line; (b) low frequency distortion; (c) low and high frequency distortion.}
    \label{fig:crack_roughness}
\end{figure}

Obviously, real-world cracks could vary widely in terms of shape, extension and severity. To encompass such variability, each meta-annotation is associated to a set of parameters that will govern the final aspect of the corresponding cracks. Among them, the fundamental parameters are:
\begin{itemize}
    \item \textit{Length}, the percentage range of the line to be covered.
    \item \textit{Roughness}, the range of low and high frequency perturbations of the line.
    \item \textit{Thickness}, the range of crack width.
    \item \textit{Depth}, the range of crack inner depth.
    \item \textit{Probability of appearance}.
\end{itemize}

\begin{figure}[tb]
    \centering
    \includegraphics[width=0.9\linewidth]{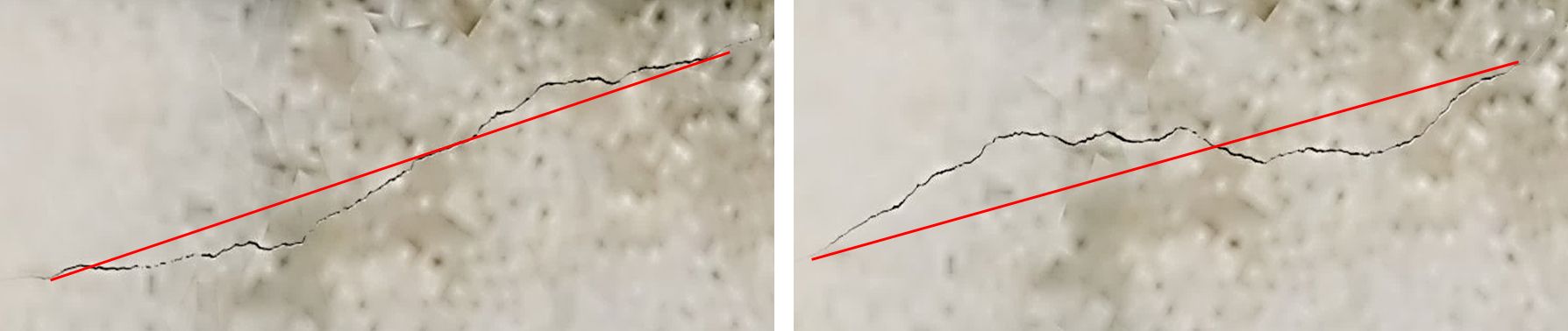}
    \caption{Different crack instances generated from the same meta-annotation (the red line).}
    \label{fig:generation}
\end{figure}

By randomly picking values within the specified parametric ranges, it is possible to generate crack instances having a wide variety of patterns and aspects. For example, Fig. \ref{fig:crack_roughness} shows the possible effects of varying roughness, while Fig. \ref{fig:generation} shows different crack instances generated by the same meta-annotation.

\begin{figure}[tb]
    \centering
    \includegraphics[width=0.9\linewidth]{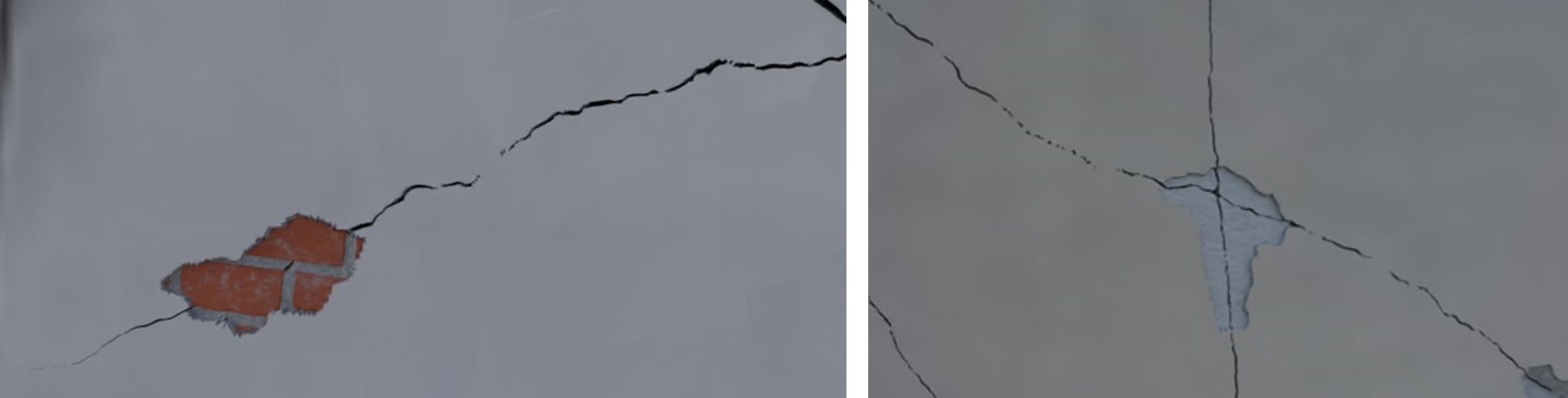}
    \caption{Example of cracks with small spallings.}
    \label{fig:crack_spalling}
\end{figure}

Real-world cracks may also cause the appearance of spalling (i.e., detachments of the external layers of a surface). Therefore, in our method, it is possible to specify the probability to generate small spallings in random positions along the crack path (Fig. \ref{fig:crack_spalling}).

\begin{figure}[htb]
    \centering
    \includegraphics[width=0.85\linewidth]{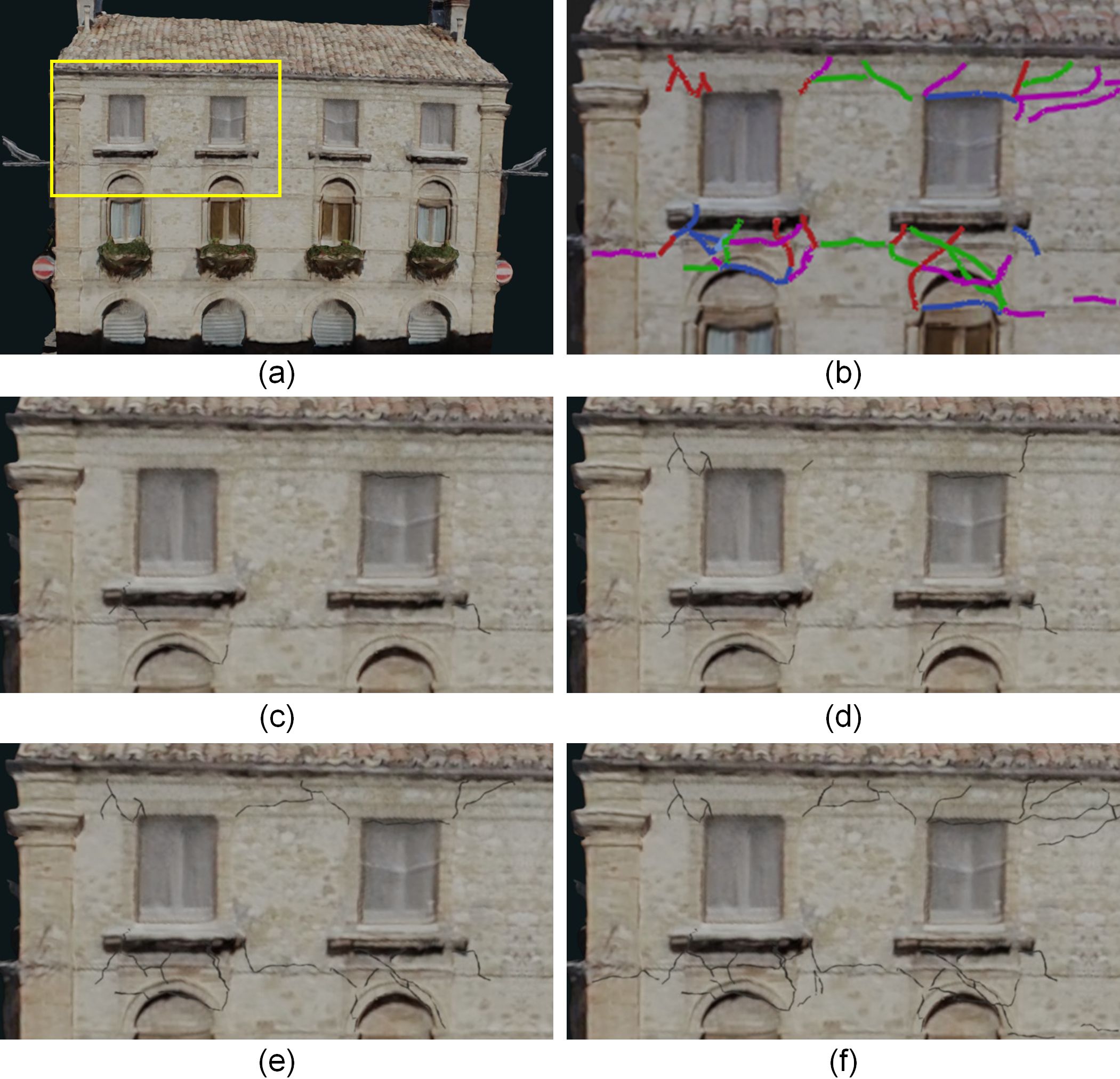}
    \caption{Example of damage evolution for a building: (a) initial 3D model with a section highlighted in yellow; (b) same section with all the generated cracks highlighted, different colors mark different groups that are enabled in sequence; (c)(d)(e)(f) renderings of the same section at increasing levels of damage severity.}
    \label{fig:damage_levels}
\end{figure}

Meta-annotations can also be aggregated to make them share a unique set of parameter values. This allows, for example, to have different groups of cracks each corresponding to the same degree of severity (e.g., a group could contain only short and thin cracks, while another may contain long and large ones). Aggregation of meta-annotations can aslo be used to simulate the actual evolution of damage on structures, as the severity conditions worsen \cite{Senaldi2020}. An example of this  is shown in Fig. \ref{fig:damage_levels}.

\subsubsection{Damage Generation}
\label{sec:damage_generation}
The generation of crack instances involves the alteration of the original 3D mesh. This operation is automatically performed by the system before the rendering process. At this stage, the straight line corresponding to each meta-annotation is converted into a 3D curve, which is altered randomly in length, roughness, thickness and depth within the parametric range of values specified. The obtained curve is then subtracted from the 3D mesh applying a boolean operation. The shape of each crack will also depend on the type of structure and surface being considered. For example, on a masonry wall, cracks will follow the spaces between the bricks. In case of spalling, since the original 3D models are limited only to the outermost surface of structures, all the internal layers will be automatically simulated by creating surfaces with ad-hoc textures and altering the 3D mesh in the surroundings of the crack line (Fig. \ref{fig:crack_spalling}).

\subsubsection{Image Rendering}
\label{sec:rendering}
Given that the objective is to generate images as similar as possible to those acquired during a UAS survey, sequences of camera movements and angles need to be specified to simulate possible flight paths. World parameters for the entire 3D model can be specified as well, for each virtual flight. To this purpose, we made use of a specific Blender add-on, called Sun Position, which simulates various lighting and meteorological conditions given a geographic location, time and date. This makes it possible to reproduce the same conditions of the day of the actual acquisition of the 3D model, thus helping to match shadows that are already present in the model itself. 

The complete setup of a rendering sequence may contain multiple flight paths with repeated generations of different levels of damage and different lighting and meteorological conditions. Once the setup is completed, the rendering process runs automatically producing a large number of images. 

\begin{figure}[htb]
    \centering
    \includegraphics[width=0.85\linewidth]{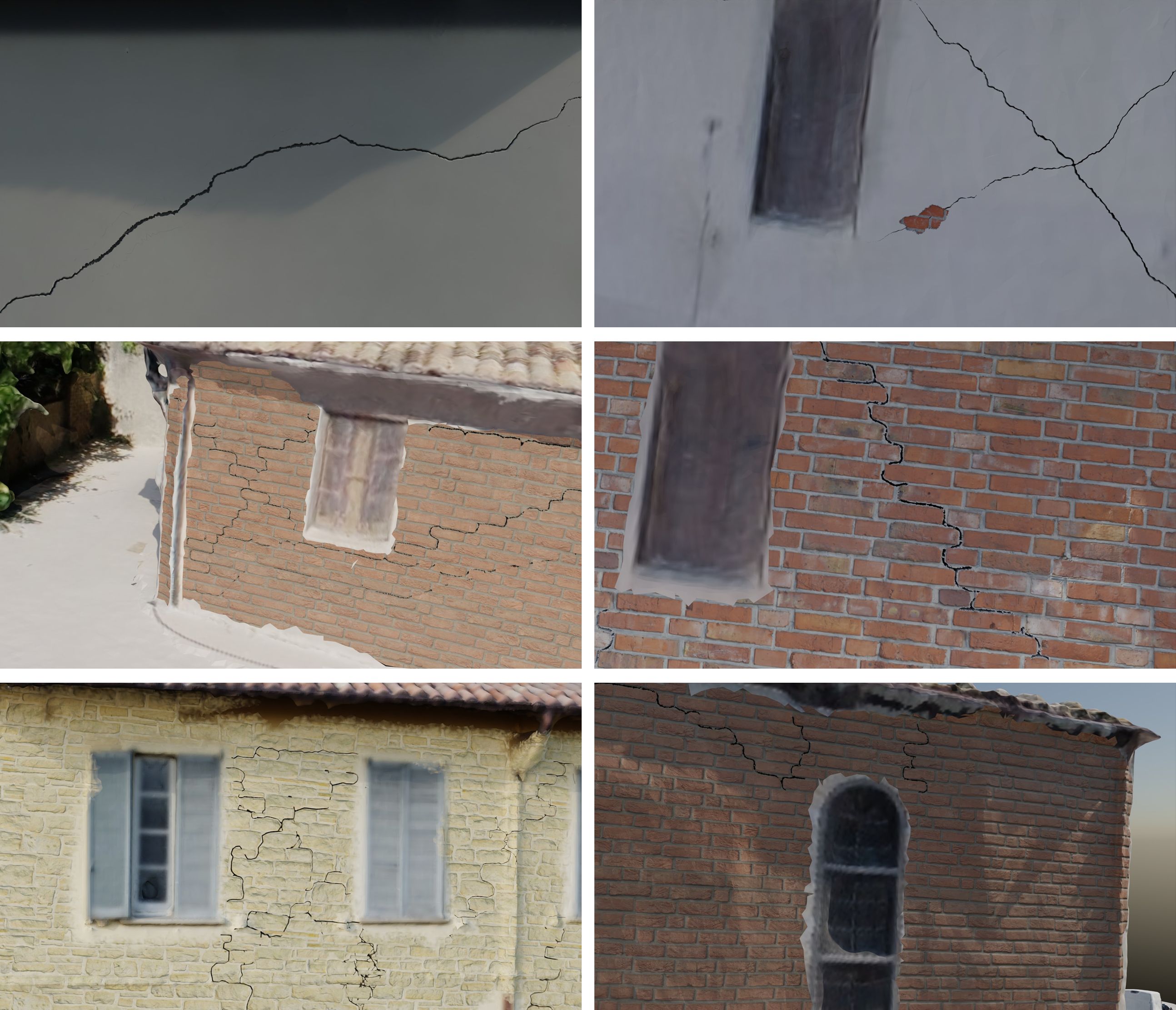}
    \caption{Examples of rendering of crack instances for various types of buildings at different distance and under different lighting and meteorological conditions.}
    \label{fig:example_buildings}
\end{figure}

\begin{figure}[htb]
    \centering
    \includegraphics[width=0.85\linewidth]{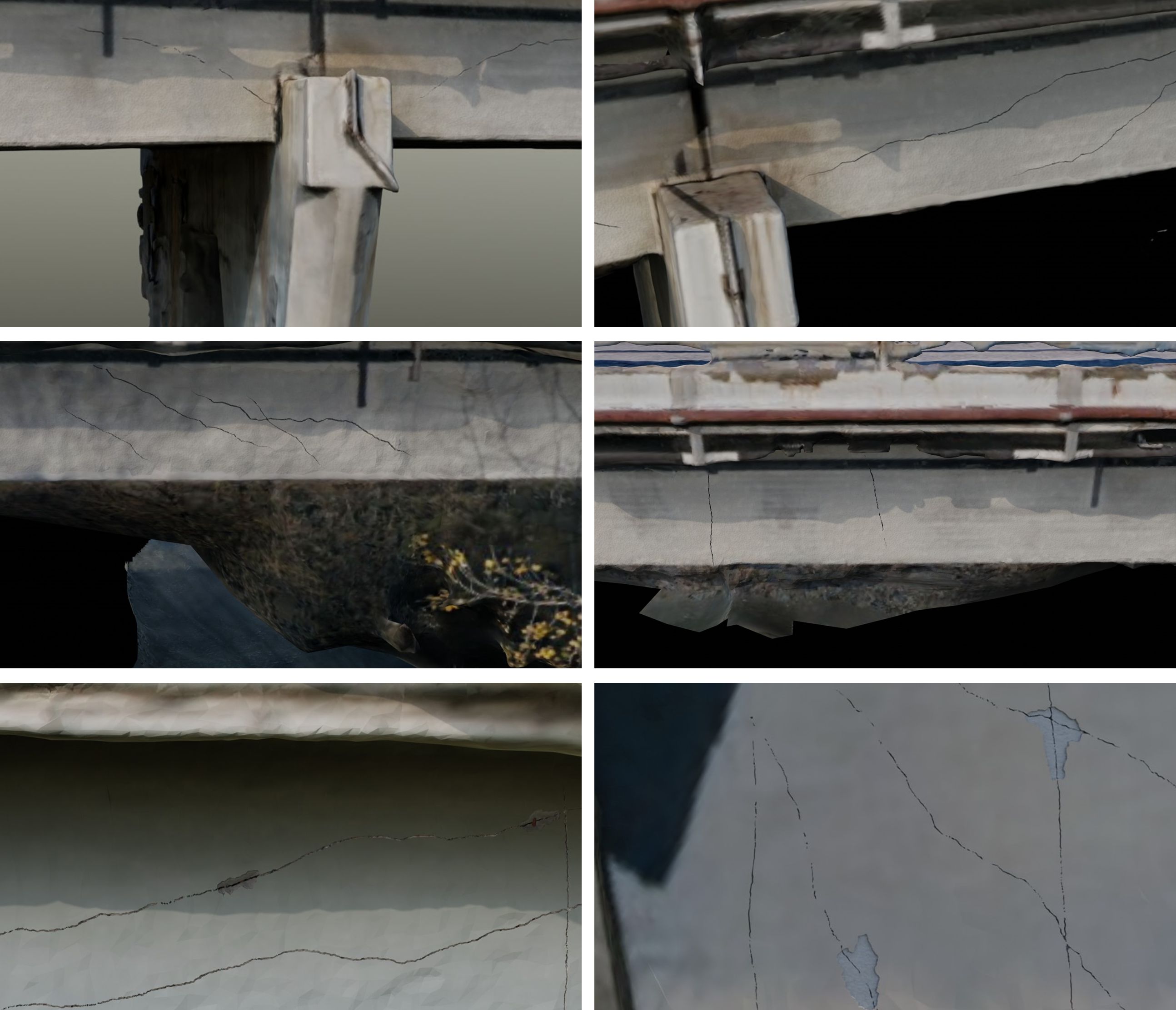}
    \caption{Examples of rendering of crack instances on bridges at different distance and under different lighting and meteorological conditions.}
    \label{fig:example_bridges}
\end{figure}

For every virtual flight, multiple frames are rendered. Figures \ref{fig:example_buildings} and \ref{fig:example_bridges} show examples of individual frames that could be generated for various types of buildings and bridges. For each frame, the system also produces a segmentation mask that describes the area corresponding to each crack. In turn, bounding box annotations are generated from segmentation masks and translated into the Pascal VOC XML format \cite{Everingham2015}. Optionally, for debugging purposes, frame copies with overlapping bounding boxes can be generated as well. 

\subsection{Tunable DCNN Training Strategy}
\label{sec:training_strategy}
As already described, the proposed meta-annotation mechanism allows to generate semi-synthetic images tuned to improve the performance of a DCNN-based crack detector. In the intended strategy, a DCNN is iteratively trained with a combination of real and semi-synthetic images. At each stage, the results obtained by the detector on the same test set of real images are examined in detail, to find out those critical cases in which crack detection is problematic. Therefore, the setup of the generation process can be altered to produce more damage instances of these critical cases. For example, if the detector frequently misses cracks on masonry walls, the process can be oriented towards the generation of more images of that kind. This operation can be performed by changing a limited number of parameter values and/or re-using existing meta-annotations. 

\section{Case Study}
\label{sec:case_study}

\subsection{Real-Image Dataset}
\label{sec:eucentre_dataset}
The IDEA (Image Database for Earthquake damage Annotation) dataset\footnote[1]{Access is granted by EUCENTRE Foundation upon reasonable request.}, created by the EUCENTRE Foundation, contains high-resolution images relating to three recent major events in Italy: L’Aquila (2009), Emilia (2012) and Central Italy (2016-2017). To the best of the authors' knowledge, to date, this is the largest available dataset of annotated images of damaged buildings and bridges acquired during post-earthquake surveys. Damage instances were manually annotated by human experts, by applying bounding boxes associated to damage class labels. 

At present, the dataset is growing thanks to the activity of the EUCENTRE Foundation. At the time of the experiments described, IDEA contained 4946 annotated images (2317 framing damaged buildings/bridges and 2629 non-damaged buildings/bridges). Overall, the crack occurrences were 2540. For the experiments reported, the IDEA dataset was split into two subsets, one for training and one for testing, containing 3952 and 994 images, respectively. Such split was performed manually to ensure the maximum possible balancing of structure types and the same proportion of images with or without damage. 

\subsection{Semi-Synthetic Dataset}
\label{sec:semisynth_dataset}
This dataset was generated incrementally with the proposed image generation method. We used as input 3D models of four building compounds and four bridges obtained via the Structure for Motion (SfM) photogrammetric technique \cite{pepe2020}. The high-resolution images needed for photogrammetry were acquired using various UAS devices (Dji Mavic 2 Pro, Air 2S, Mini 2) during multiple flights performed between 2018 and 2022. Agisoft Metashape was the photogrammetry software of choice. The produced 3D meshes were cleansed of minor defects, due to acquisition inaccuracies, using Meshlab \cite{Cignoni2008}.

This semi-synthetic dataset was extended incrementally during the training process, in keeping with the strategy described in Section \ref{sec:training_strategy}. At the end of the process, the dataset contained 33836 semi-synthetic images of buildings and bridges, 30059 damaged and 3777 non-damaged. In total, the generated cracks were more than 135K.

\subsection{Experiments}
To validate the advantages of the proposed approach, we conducted a series of comparative experiments. We trained various DCNN-based crack detectors: (i) the baseline model, using only real images from the training split of the IDEA dataset; (ii) a second model, using only the semi-synthetic dataset; (iii) the augmented model, using the same training split of the IDEA dataset used for the baseline model with augmentation of the semi-synthetic dataset. The same testing split of the IDEA dataset was used in all cases.

As DCNN architectures, we tested the YOLOv8 architecture developed by Ultralytics \cite{YOLOv8}, YOLOv9 \cite{YOLOv9} and YOLOv10 \cite{YOLOv10}. More precisely, for each of them, we used the small model, 1088 image size, and 50 epochs for training. The choice of the small model was dictated by the requirement of having the smallest possible computational footprint for the detector, to make it usable on field with limited computational capabilities. For validation reliability, all hyperparameters were tuned separately for each detector. 

\section{Experimental Results}
\label{sec:results}
At first, we trained the baseline model. After some initial tests, the need to apply a correction to the bounding boxes of the IDEA dataset emerged. In fact, since cracks have imprecise boundaries, human experts may draw bounding boxes that do not cover them completely. To compensate for possible inaccuracies of this kind, we expanded the bounding boxes in all directions. We tested expansions from 1-pixel to 7-pixel wide, and we attained the best outcomes with a 3-pixel expansion. Table \ref{tab:yolos} shows the results obtained with the three YOLO architectures considered, measured in terms of Average Precision (AP) with the threshold of Intersection-over-Union (IoU) set at 0.5. As can be noticed, the best result was achieved with YOLOv8. Given the clear difference in performance, in the subsequent experiments we report only the results of YOLOv8. 

\begin{table}[tb]
    \centering
    \caption{Comparison among different YOLO architectures, trained and tested on the IDEA dataset.}
    \begin{tabular}{p{3cm}c}
        \toprule
        Model  & AP@05 \\
        \midrule
        YOLOv8 small  & 0.382 \\
        YOLOv9 small  & 0.355 \\
        YOLOv10 small & 0.311 \\
          \bottomrule
    \end{tabular}

    \label{tab:yolos}
\end{table}

\begin{figure}[!htb]
    \centering
    \includegraphics[width=0.88\linewidth]{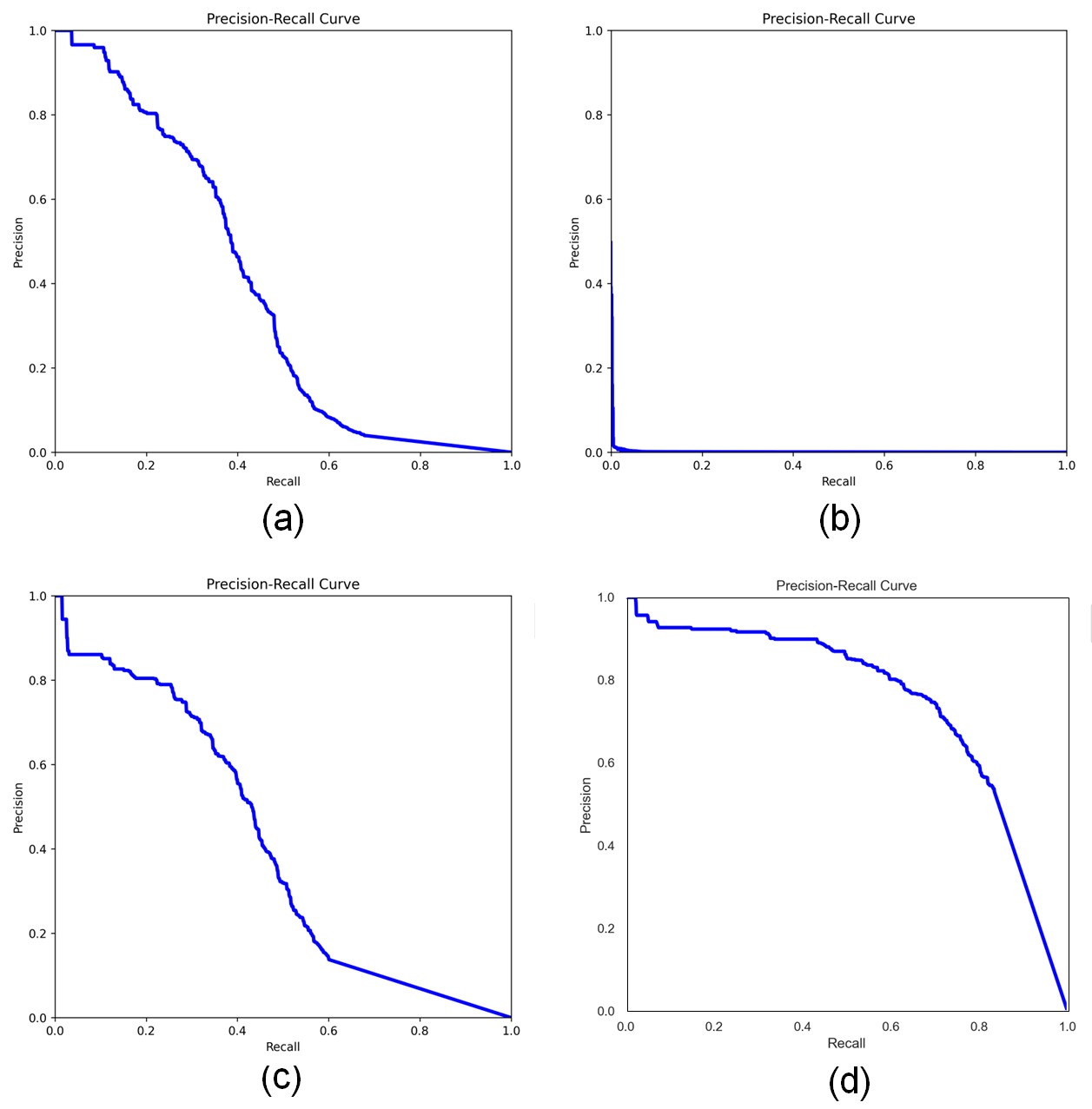}
    \caption{Best Precision-Recall (PR) curves obtained by YOLOv8 with different training strategies, in all cases the test set contains only real images: (a) baseline model, trained on real images only; (b) model trained on semi-synthetic image only; (c) augmented model, trained on real images plus semi-synthetic images augmentation; (d) augmented model using the Many-to-Many metrics instead of the standard metrics.}
    \label{fig:PR_curves}
\end{figure}

\begin{figure}[htb]
    \centering
    \includegraphics[width=\linewidth]{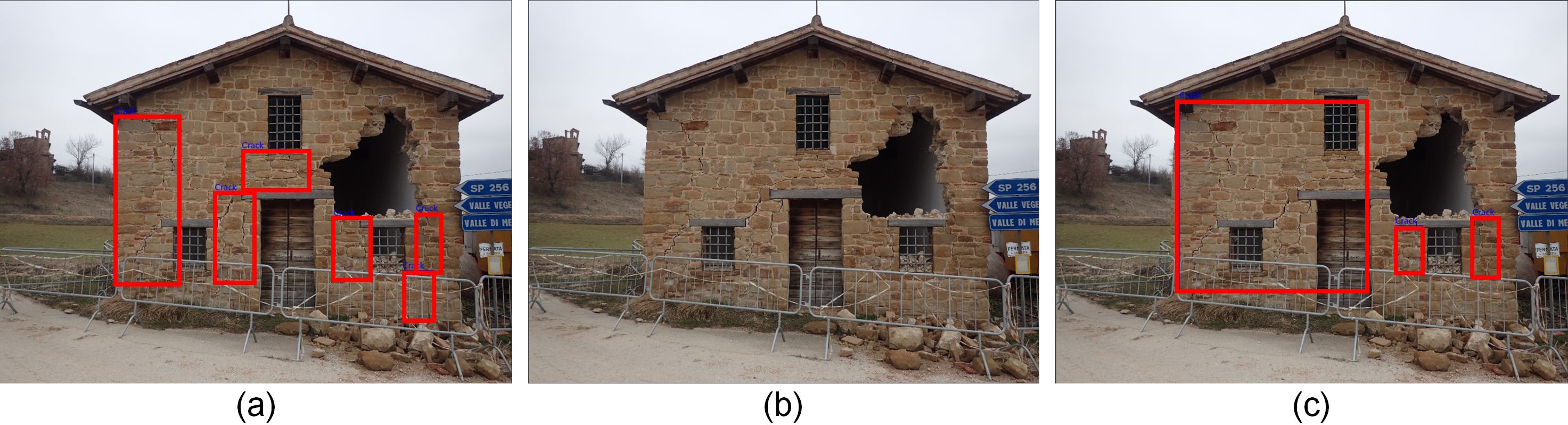}
    \caption{Example of crack detection on a test image of the IDEA dataset: (a) ground truth; (b) prediction of the baseline model; (c) prediction of the augmented model.}
    \label{fig:example_detection}
\end{figure}

Figure \ref{fig:PR_curves}(a) shows the Precision-Recall (PR) curve for the baseline model. 
Figure \ref{fig:PR_curves}(b) shows instead the PR curve obtained with a model trained with semi-synthetic images only. In this case, the results are indeed poor, meaning that semi-synthetic images could not be used as a pure replacement for real-world images. Figure \ref{fig:PR_curves}(c) shows the best PR curve obtained with the augmented model, which was trained using semi-synthetic images as augmentation of the real ones. Given that semi-synthetic images largely outnumber the real ones, we applied a rebalancing strategy to avoid biasing the detector towards semi-synthetic data. More precisely, we oversampled with a ration 1:6 the images from the IDEA training split to match the amount of semi-synthetic images. Comparing Fig. \ref{fig:PR_curves}(a) and (c), we can notice an improvement in performance of the augmented model with respect to the baseline, as the AP@0.5 grows from 0.382 to 0.407. Figure \ref{fig:example_detection} shows an example of the enhanced detection capabilities of the augmented model with respect to the baseline. 

In the same example, Fig. \ref{fig:example_detection}(a)(c), we can notice that bounding boxes vary significantly in both number and size between ground truth and prediction. Nonetheless, most of the cracks are detected anyhow. This is a common outcome in the considered scenario. In fact, unlike standard target objects like cars or pedestrians, cracks may have ambiguous shapes and boundaries, which could be annotated and detected in several different ways. This phenomenon may lead to the underestimation of the actual performance of a detector, since standard IoU metrics are typically applied to the best one-to-one match between ground-truth and predicted bounding boxes only. To better deal with this, the Many-to-Many metrics were proposed in \cite{Dondi2023}. These new metrics assume that the same Ground Truth Box (GTB) can be matched by multiple Prediction Boxes (PBs), and vice versa. In the Many-to-Many approach, Precision and Recall are computed using two different variants of the IoU metric: \textit{Intersection over Prediction} (IoP, see Eq. \ref{eq:IoP}) is used to evaluate Precision, while \textit{Intersection over Ground Truth} (IoG, see Eq. \ref{eq:IoG}) is used to evaluate Recall. 

\begin{equation}
\label{eq:IoP}
    IoP=\frac{areas \ of \ overlap}{area \ of \ PB}
\end{equation}

\begin{equation}
\label{eq:IoG}
    IoG=\frac{areas \ of \ overlap}{area \ of \ GTB}
\end{equation}

Figure \ref{fig:PR_curves}(d) shows the PR curve for the augmented model recomputed using the Many-to-Many metrics, with thresholds for both IoP and IoG at 0.5. We can notice a significant increase in the AP@0.5, from 0.407 to 0.749, as well as a smoother curve. In reference to the example in Fig. \ref{fig:example_detection}, the Many-to-Many metrics appear to yield a better estimation of the actual performance of the crack detector.

\section{Conclusions}
\label{sec:conclusions}
In this work we have presented a method for generating a large number of semi-synthetic images of damaged buildings and bridges, starting from 3D models of real-world structures. The generation process is guided by parametric meta-annotations, which offer significant flexibility and control over the outcome. Specifically, they allow to tune the image generation process to match the gaps in coverage that could be present in a real-world dataset. The semi-synthetic images thus obtained can be used as an effective data augmentation of real images during the training process of a DCNN-based detector. 

Experiments conducted on a real-world dataset demonstrate that the proposed tunable training strategy improves the performance of a crack detector. In this perspective, this work contributes to implementing more effective screening procedures during post-earthquake surveys, with the final goal of speeding up the assessment of the impacted areas. 

Future steps will involve the extension of the semi-synthetic image generation process to other  classes of damage, such as spalling and exposed rebars. We also plan to move the analysis from single images to full videos acquired by UAS. Video analysis will allow us to track damage instances and to employ temporal filtering to further refine the detection.

\section*{Acknowledgments}
The current work has been carried out under the financial support of the project TeamAware (European Union’s Horizon 2020 research and innovation program, grant agreement No 101019808) and of the Italian Civil Protection, within the framework of the Executive Project 2022-2023 and 2024–2026.

\bibliographystyle{splncs04}
\bibliography{mybib}
\end{document}